\documentclass{article}
\usepackage{ijcai24}

\usepackage{times}
\usepackage{soul}
\usepackage{url}
\usepackage[hidelinks]{hyperref}
\usepackage[utf8]{inputenc}
\usepackage[small]{caption}
\usepackage{graphicx}
\usepackage{amsmath}
\usepackage{amsthm}
\usepackage{booktabs}
\usepackage{algorithm}
\usepackage{algorithmic}
\usepackage[switch]{lineno}

\usepackage{multirow}
\usepackage{makecell}
\usepackage{xcolor}
\usepackage{xspace}

\newcommand{\mhy}[1]{\textcolor[rgb]{0,0,0}{#1}}

\newcommand\ie{\textit{i.e.,} }

\title{\mhy{Apprenticeship-Inspired Elegance: Synergistic Knowledge Distillation\\Empowers Spiking Neural Networks for Efficient Single-Eye Emotion Recognition}}

\author{
Yang Wang$^{1,2}$
\and
Haiyang Mei$^{2,1}$\and
Qirui Bao$^{1}$\and
Ziqi Wei$^{3}$\and
Mike Zheng Shou$^{2}$\and 
Haizhou Li$^{5,2}$\and \\
Bo Dong$^{4}$\And
Xin Yang$^{1}$\thanks{Corresponding author.}\\
\affiliations
$^1$Key Laboratory of Social Computing
and Cognitive Intelligence, Dalian University of Technology \\
$^2$National Univerisity of Singapore \\
$^3$Institute of Automation, Chinese
Academy of Sciences \\
$^4$Independent Researcher \\
$^5$The Chinese University of Hong Kong, Shenzhen (CUHK-Shenzhen)\\
\emails
\{yangwang06,1612000589\}@mail.dlut.edu.cn,
haiyang.mei@outlook.com,
ziqi.wei@ia.ac.cn,
\{mike.zheng.shou,dongshuhao12\}@gmail.com,
haizhouli@cuhk.edu.cn,
xinyang@dlut.edu.cn
}

\begin{document}

\maketitle

\begin{abstract}
We introduce a novel multimodality synergistic knowledge distillation scheme tailored for efficient single-eye motion recognition tasks. This method allows a lightweight, unimodal student spiking neural network (SNN) to extract rich knowledge from an event-frame multimodal teacher network. The core strength of this approach is its ability to utilize the ample, coarser temporal cues found in conventional frames for effective emotion recognition. Consequently, our method adeptly interprets both temporal and spatial information from the conventional frame domain, eliminating the need for specialized sensing devices, \textit{e.g.}, event-based camera. The effectiveness of our approach is thoroughly demonstrated using both existing and our compiled single-eye emotion recognition datasets, achieving unparalleled performance in accuracy and efficiency over existing state-of-the-art methods. 

\end{abstract}

\section{Introduction}
\label{sec:intro}
Real-time emotion recognition is pivotal in enhancing human-centered interactive experiences, such as virtual reality (VR) and augmented reality (AR) applications \cite{picard2003affective,zhang2023blink}. This technology's proficiency in precisely decoding users' emotional states can greatly elevate the VR/AR experience. More significantly, it facilitates the personalization of these experiences to cater to the distinct emotional requirements of each user, thereby offering uniquely immersive experiences and boosting engagement. 

In the context of VR/AR, the devices are typically affixed to a user's face, which inherently accommodates the variances in performance that may arise from different head positions. While beneficial for head pose accommodation, this placement presents a significant challenge: the majority of the facial area is obscured by the device, diminishing the effectiveness of traditional emotion recognition methods that rely on facial action units. To counteract this limitation, the focus is shifting towards eye-based emotion recognition techniques~\cite{hickson2019eyemotion,wu2020emo}. Yet, these methods often depend on personalized initialization or necessitate capturing the peak phase of an emotion~\cite{hickson2019eyemotion}, which can limit their practicality.

\begin{figure}[t]
  \centering
  \includegraphics[width=1\linewidth]{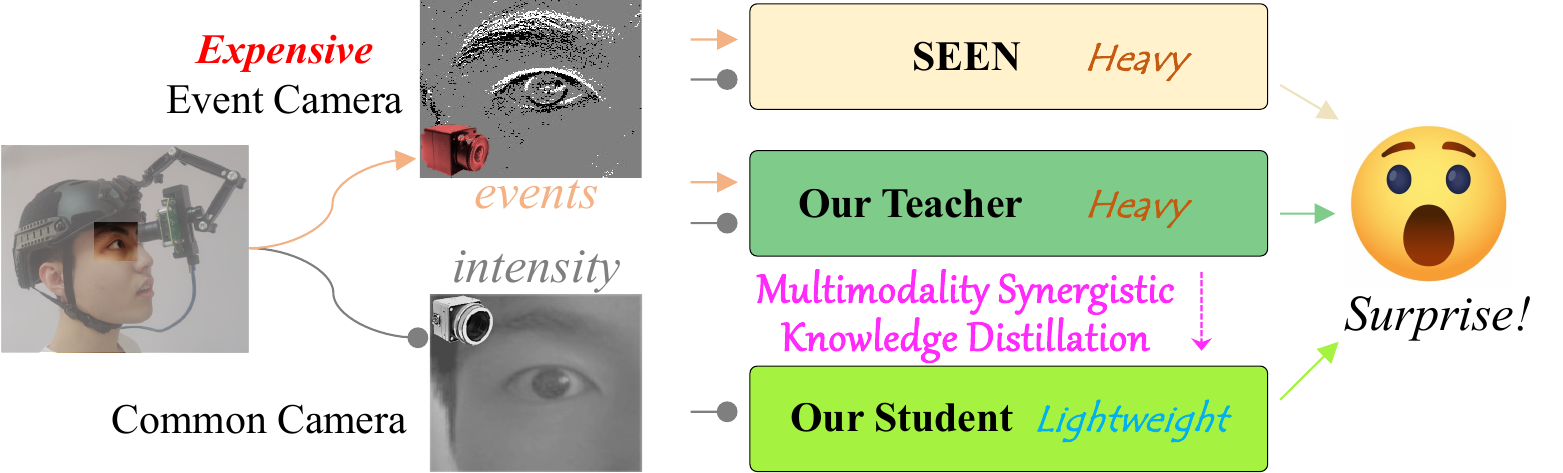} 
  \caption{\mhy{Surpassing the state-of-the-art SEEN in real-time single-eye emotion recognition (SER), our method achieves lightweight inference by requiring solely intensity frames, obviating the need for data from expensive event cameras. This is facilitated by our novel synergistic knowledge distillation strategy, enabling real-time and accurate SER on resource-constrained devices for the first time.}}
  \label{fig:teaser}
\end{figure}

Recently, an event-based single-eye-based approach \cite{zhang2023blink} capitalizes on both temporal and spatial cues, enhancing accuracy in emotion recognition without needing personalization or capturing peak emotional phases. It proves effective under various real-world lighting conditions, including low-light and high-dynamic-range environments. The success of this approach is underpinned by the advantages of event-based cameras, which offer a higher dynamic range (140 dB vs. 80 dB in traditional cameras) and a significantly finer temporal resolution (about 0.001 ms, in contrast to 10 ms in conventional frame-based cameras). However, despite these advancements, event-based cameras are still nascent, especially when compared to the more established conventional frame-based camera technologies. This relative immaturity translates to higher costs, making the widespread adoption of event-based cameras in VR/AR devices not currently a cost-effective solution. 

In our study, we developed an \textit{apprenticeship-learning-based} approach to address the aforementioned limitation. Our method efficiently learns from both event and frame domains but uses only conventional frames for inference; see~\autoref{fig:teaser}. In particular, our approach employs a knowledge distillation process, where a teacher network, trained on multimodal data (event and frame), transfers its insights to a student network. This student network is then adept at harnessing spatial and temporal cues from conventional frames. The key to this method's success lies in leveraging the sufficient coarser temporal cues present in conventional frames for effective emotion recognition, eliminating the need for more expensive event-based cameras. The efficacy of our distillation scheme is reinforced by two novel consistency losses we developed: hit consistency and temporal consistency. Hit consistency ensures a match in the correct prediction distribution between the teacher and student networks. Temporal consistency, on the other hand, encompasses all temporal predictions, considering both correct and incorrect ones. We extensively validate the efficacy of our approach on both SEE dataset~\cite{zhang2023blink} and our Diverse Single-eye Event-based Emotion (DSEE) datasets, demonstrating the best performance among all competing state-of-the-art methods.
In summary, our contributions include:
\begin{itemize}
    \item Developing a novel framework for an unimodal student network to distill knowledge from an event-enhanced multimodal teacher network;
    \item Creating a new large-scale dataset with both frame and event for advancing eye-based emotion recognition;
    \item Achieving significant improvements in accuracy and efficiency over competing state-of-the-art methods.
\end{itemize}

\section{Related Work}
\label{sec:related_work}

\mhy{\textbf{Facial-based emotion recognition} has received significant attention in recent years given its diverse and practical applications in the field of security, health, communication, etc. A number of facial emotion recognition datasets have been developed to facilitate the research and development of this field, such as CK+~\cite{lucey2010extended}, MUG~\cite{aifanti2010mug}, MMI~\cite{pantic2005web}, Oulu-CASIA~\cite{zhao2011facial}, and ADFES~\cite{van2011moving}.
The majority of previous research focuses on analyzing the entire face, with various techniques introduced for effective facial feature learning~\cite{xue2021transfer,ruan2021feature}, addressing uncertainties in facial expression data~\cite{zhang2021relative}, handling partial occlusions~\cite{georgescu2019recognizing,houshmand2020facial}, and utilizing temporal cues~\cite{sanchez2021affective,deng2020mimamo}.
However, in many practical scenarios, it is not always feasible to observe the entire face, which triggers growing interest in identifying emotions based solely on information from the eye area.}

\mhy{\textbf{Eye-based emotion recognition} is a branch of occluded facial emotion recognition. Years of dedicated investigation have yielded substantial progress, demonstrating the potential of this avenue for enhancing the accuracy and robustness of emotion detection. \cite{hickson2019eyemotion} utilized images of both eyes captured with an infrared gaze-tracking camera within a virtual reality headset to infer emotional expressions while \cite{wu2020emo} relied on infrared single-eye observations to address camera synchronization and data bandwidth issues when monitoring both eyes. Both constructed systems necessitate a personalized initialization procedure: the former requires a personalized neutral image while the latter needs a reference feature vector of each emotion. The requirement for a personalized setup renders these systems intrusive and non-transparent to the user, potentially raising privacy concerns. Additionally, neither system incorporates temporal cues, which are crucial for robust emotion recognition~\cite{sanchez2021affective}. Most recently, \cite{zhang2023blink} proposed a new Single-eye Event-based Emotion dataset (SEE) and a real-time emotion recognition method SEEN that integrates event and intensity cues and achieves enhanced emotion recognition. Our method distinguishes itself from SEEN by distilling enriched knowledge from both event and intensity modalities to a lightweight SNN model during training. This eliminates the requirement for the expensive event camera during the inference stage while enabling high-speed and low-energy-cost emotion recognition.}

\begin{figure*}[t]
  \centering
  \includegraphics[width=1\textwidth]{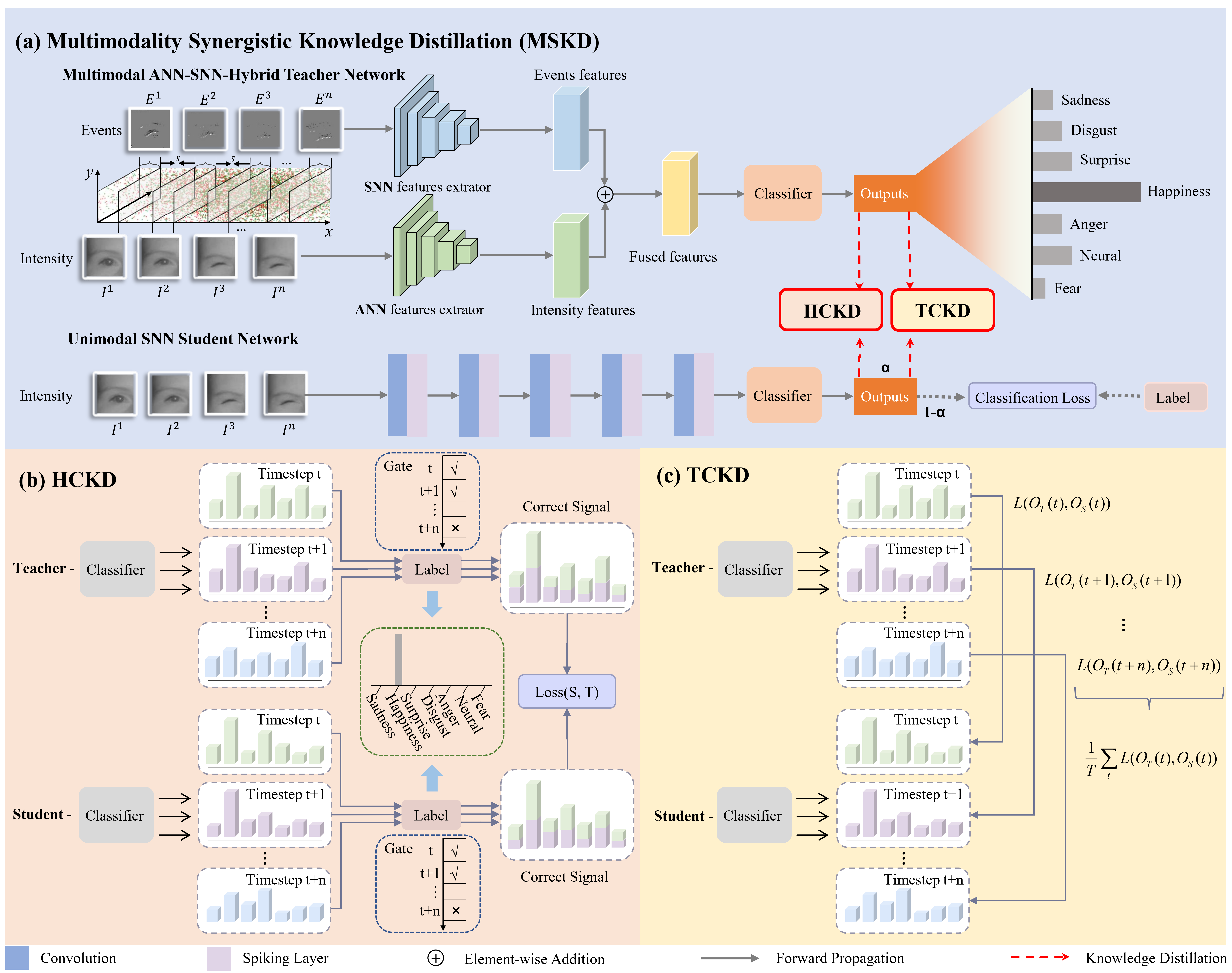} 
  \caption{\mhy{Overview of our proposed multimodality synergistic knowledge distill (MSKD) framework (a) which consists of a multimodal input ANN-SNN-hybrid teacher network (top) and an unimodal input SNN student network (bottom), as well as two synergistic knowledge distill loss items: (b) hit consistency loss and (c) temporal consistency loss.}}
  
  \label{fig:pipeline}
\end{figure*}

\mhy{\textbf{SNN-based knowledge distillation} has emerged as a promising approach to address the challenges of training deep SNNs directly with a loss function, which is hindered by the non-differentiable nature of spiking signals~\cite{wei2024event,zhang2021rectified,liu2020unsupervised,liu2020effective}. This technique unlocks the potential of deep SNNs for efficient yet accurate inference. Related works can be broadly categorized into two categories: those distilling knowledge~\cite{ji2023teachers} from differently-structured and pre-trained artificial neural networks (ANNs)~\cite{xu2023constructing,takuya2021training} or SNNs~\cite{xu2023biologically,kushawaha2021distilling}, and those distilling knowledge from itself~\cite{dong2023temporal,deng2022temporal}. Unlike the prior studies, our proposed multimodality knowledge distillation strategy achieves knowledge transfer from a multimodal input network to an unimodal input network.}

\section{Methodology}

\mhy{While existing research has demonstrated the effectiveness of multimodal networks for real-time emotion recognition (\textit{e.g.}, \cite{zhang2023blink}), their inherent complexity and need for multimodal data raise the question: \textit{can a lightweight unimodal network achieve comparable performance?} In this work, we make the first investigation into this question by proposing a novel method leveraging knowledge distillation.}

\mhy{As illustrated in \autoref{fig:pipeline}(a), we first train a cumbersome teacher network that takes as input the events data and intensity frames. The teacher network uses an SNN-based SEW-Resnet-18~\cite{fang2021deep} and a CNN-based Resnet-18~\cite{he2016deep} to extract event features and intensity features, respectively, based on which to perform emotion recognition via features fusion and fully connected classifier. We then construct a lightweight SNN-based student network that operates solely on intensity frames and consists of five consecutive feature extraction layers and a classifier, optimizing it using a classification loss (\autoref{sec:Cls_loss}) alongside two synergistic knowledge distillation losses that ensure prediction distribution harmony with the teacher network at both granular (hit consistency, \autoref{sec:hckd}) and comprehensive (temporal consistency, \autoref{sec:tckd}) levels. Formally, the loss function is defined as:
\begin{equation}\label{eq:all_loss}
\mathcal{L}_\text{total} = (1 - \alpha) * \mathcal{L}_{\text{Cls}} + \alpha  *\mathcal{L}_\text{HCKD} + \alpha *\mathcal{L}_\text{TCKD},
\end{equation}
where $\mathcal{L}_{\text{Cls}}$ is the classification loss to enforce the output at each timestep close to the true label; $\mathcal{L}_\text{HCKD}$ and $\mathcal{L}_\text{TCKD}$ are the hit consistency and temporal consistency knowledge distillation losses to improve the prediction distribution matching with the teacher network at the individual correctly predicted timestamp and each of all timestamps, respectively; and $\alpha$ is the weighting parameter to balance the classification loss and distillation losses, which is initialized as 0.5 and increased by 0.1 after every 30 training epochs.}


\subsection{Classification Loss}\label{sec:Cls_loss}
\mhy{Typically, it's not easy to efficiently train deep SNNs due to the non-differentiability of its activation function \cite{ding2022biologically,zhang2022spiking,wang2023event}, which disables the widely used gradient descent approaches for traditional ANNs. Although the adoption of surrogate gradient (SG) formally allows for the back-propagation of losses, the discrete spiking mechanism differentiates the loss landscape of SNNs from that of ANNs, failing the SG methods to achieve desirable accuracy. To alleviate this, we follow \cite{deng2022temporal} to adopt the temporal efficient training (TET) approach to compensate for the loss of momentum in the gradient descent with SG so that the training process can converge into flatter minima with better generalizability. Equipped with TET, our classification cross-entropy loss can be defined as:
\begin{equation}
\label{eq:Cls_loss}
\mathcal{L}_{\text{Cls}} = \frac{1}{T}\sum_{t=1}^T \mathcal{L}_{CE} (O_{stu}(t), y),
\end{equation}
where $O_{stu}(t)$ represents the pre-synaptic input of the student classifier at the $t$-\textit{th} timestep; $\mathcal{L}_{CE}$ denotes the cross-entropy loss; $T$ is the total timesteps; and $y$ indicates the ground truth one-hot vector. Different from \cite{zhang2023blink} that directly optimizes the integrated potential, our classification loss optimizes every moment’s pre-synaptic inputs, which helps the network have more robust time scalability.}


\subsection{Hit Consistency Knowledge Distillation}\label{sec:hckd}
\mhy{Despite with lightweight architecture, the student network is expected to recognize emotions as correctly as the cumbersome teacher network. This inspires our hit consistency knowledge distillation (HCKD) loss (\autoref{fig:pipeline}(b)) which penalizes the distribution difference between teacher and student networks \textit{at the correctly-predicted timestep}. Formally,
\begin{align}\label{eq:hckd}
\mathcal{L}_\text{HCKD} &= \mathcal{L}_{\text{MSE}} ( S_{stu}, S_{tea}), \\
S_{stu} &= \frac{1}{C_{stu}}\sum_{c_{stu}=1}^{C_{stu}} O_{stu}(c_{stu}), \\
S_{tea} &= \frac{1}{C_{tea}}\sum_{c_{tea}=1}^{C_{tea}} O_{tea}(c_{tea}),
\end{align}
where ${L}_{\text{MSE}}$ measures the mean squared error between student correctly-predicted signal $S_{stu}$ and teacher correctly-predicted signal $S_{tea}$. $S_{stu}$/$S_{tea}$ is obtained by averaging $C_{stu}$/$C_{tea}$ student/teacher prediction distributions $O_{stu}(c_{stu})$/$O_{tea}(c_{tea})$ at the correctly-predicted timestep $c_{stu}$/$c_{tea}$. When $C_{stu}$/$C_{tea}$ equals to zero, we assign $1/N_c$ as its value ($N_c$ is the total number of emotion categories).}



\subsection{Temporal Consistency Knowledge Distillation}\label{sec:tckd}
\mhy{The HCKD emphasizes the harmony of averaged distributions for all correctly-predict timesteps between student and teacher networks, which could help the student network approach the teacher network in terms of the \textit{\textbf{overall}} correct predictions. However, HCKD does not consider the temporal consistency between student and teacher networks at each timestep, ignoring the distillation of rich knowledge embedded in the temporal patterns and dynamics. To address this limitation, we introduce a temporal consistency knowledge distillation (TCKD) loss (\autoref{fig:pipeline}(c)), which enforces the student network to learn temporal patterns and dynamics from the teacher by quantifying the discrepancy in temporal dynamics, via computing the mean squared error between their prediction distributions \textit{\textbf{at each timestep}}. Formally,
\begin{align}\label{eq:tckd}
\mathcal{L}_\text{TCKD} = \frac{1}{T}\sum_{t=1}^T \mathcal{L}_{\text{MSE}} ( O_{stu}(t), O_{tea}(t)),
\end{align}
where $T$ is the number of all timesteps. Finally, by combining HCKD loss and TCKD loss, we form a new and powerful synergistic knowledge distillation strategy to empower a lightweight unimodal student network for efficient SER.}

\section{Dataset}\label{sec:data}

\begin{figure}[t]
    \centering
        \includegraphics[width=1.0\linewidth]{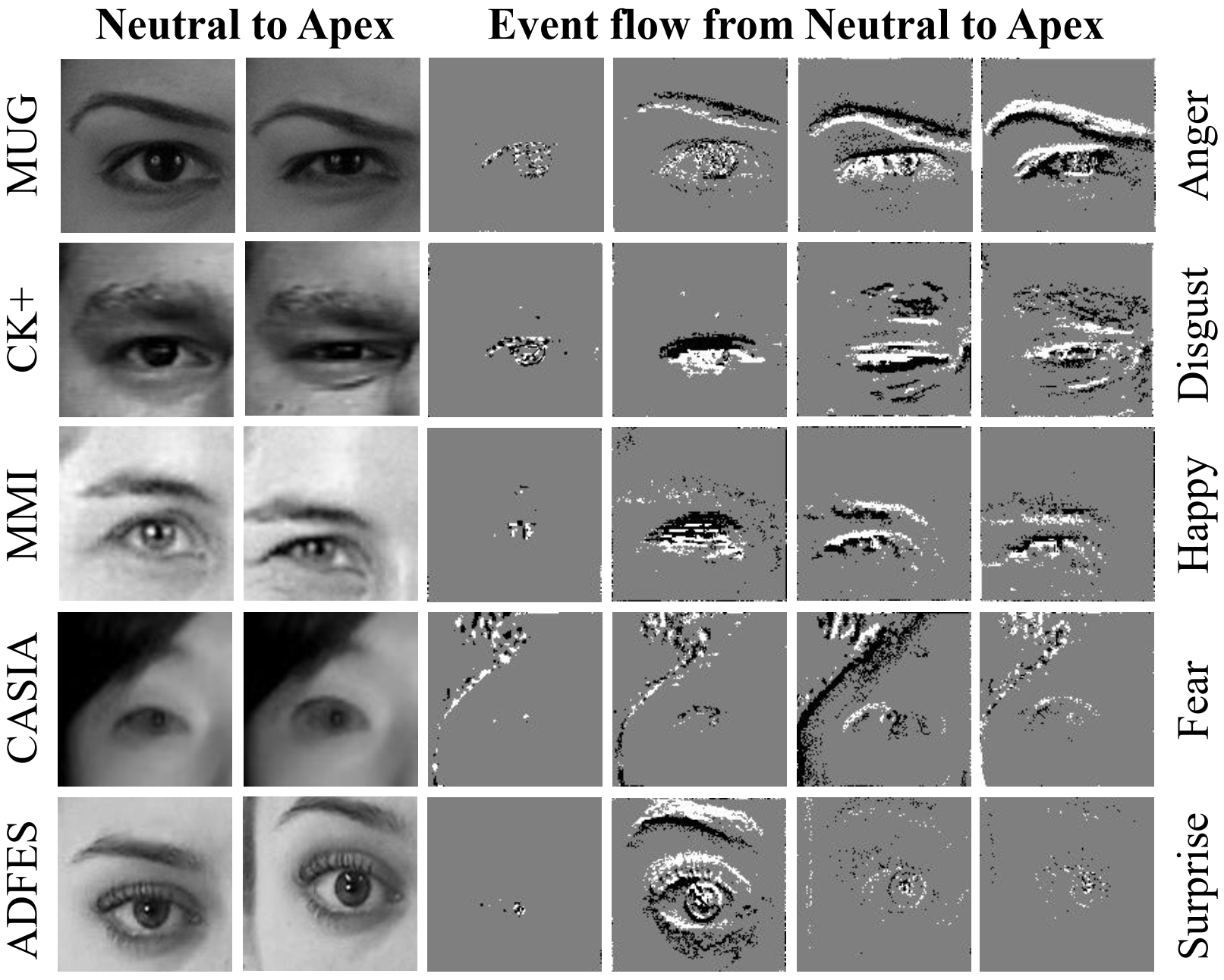}
    \caption{\mhy{Examples from our DSEE dataset.}}
    \label{fig:data_1}
\end{figure}

\begin{table*}[tb]
	\centering
    \small
    \setlength{\tabcolsep}{4pt}
		\begin{tabular}{c|cccccccccc}
		 \toprule
			Datasets & Sequences & Frames & Subjects& Emotion& Age range & Race number & Intensity & Events & Real & Synthetic \\
	   \midrule
			MUG& 983& 70596& 52 & 7& 20-35&1&$\surd$& $\surd$&$\times$& $\surd$ \\
	
			CK+&327 & 5876&118&7& 18-50&3&$\surd$& $\surd$&$\times$& $\surd$ \\
		
			MMI&126&11099&21&6& 20-32&3&$\surd$& $\surd$&$\times$& $\surd$\\

			Oulu-CASIA&1440&31995&80&6&23-58&1&$\surd$& $\surd$&$\times$& $\surd$\\
 
			Adfaces&216&32350&22&\textbf{10}&18-25&2&$\surd$& $\surd$&$\times$& $\surd$\\
               
            SEE&3224&175185&113&7&19-28&1&$\surd$& $\surd$&$\surd$&$\times$\\
            \textbf{DSEE} (Ours)&\textbf{6235}&\textbf{317447}&\textbf{394}&7&\textbf{18-58}&\textbf{6}&$\surd$& $\surd$&$\surd$& $\surd$\\
        \bottomrule
		\end{tabular}
  \caption{\mhy{Comparison among event-based datasets for emotion recognition.}}
		\label{tab_eff}
\end{table*}

\mhy{The scarcity of publicly available datasets is a major challenge in eye-based emotion recognition research. Early related datasets include the active infrared lighting/camera datasets Eyemotion \cite{hickson2019eyemotion} (both eyes) and EMO \cite{wu2020emo} (single eye). Recently, \cite{zhang2023blink} collected a Single-eye Event-based Emotion (SEE) dataset for event-enhanced emotion recognition. SEE contains data from 111 volunteers captured with a DAVIS346 event-based camera placed in front of the right eye and mounted on a helmet. Despite the pioneering effort to introduce events modality for improving recognition accuracy, SEE is highly imbalanced, with a majority of samples from challenging scenes and a minority of samples from normal scenes. In addition, SEE has limited example diversity in terms of participants' age, gender, race, etc. These two shortcomings significantly impair the generalization ability of models trained on this dataset.}


\mhy{To address the above limitations, we introduce a Diverse Single-eye Event-based Emotion (DSEE) dataset. DSEE contains intensity video frames and corresponding real/synthetic events as well as a ground truth emotion label. To the best of our knowledge, DSEE is currently the largest single-eye event-based emotion benchmark (kindly see \autoref{tab_eff} for a summary and \autoref{fig:data_1} for representative examples).}


\subsection{Protocols for Data Acquisition}
\begin{figure}
    \centering
        \includegraphics[width=1.0\linewidth]{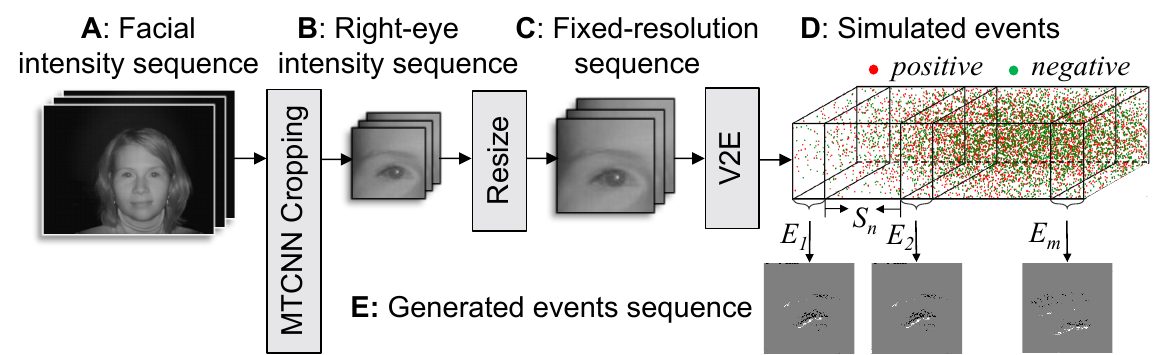}
    \caption{\mhy{Illustration of the single-eye events data synthesis process.}}
    \label{fig:syn_process}
\end{figure}

\mhy{Besides including SEE \cite{zhang2023blink} as a subset, we ensure a wide diversity and broad coverage of our DSEE by inheriting and further processing existing facial emotion recognition video datasets including CK+~\cite{lucey2010extended}, MUG~\cite{aifanti2010mug}, MMI~\cite{pantic2005web}, Oulu-CASIA~\cite{zhao2011facial}, and ADFES~\cite{van2011moving}. As illustrate in \autoref{fig:syn_process}, we first use a multi-task cascaded convolutional network (MTCNN) \cite{zhang2016joint} to locate and crop the right eye regions from the given facial sequence. Then we resize the cropped region to a fixed resolution (\ie, 128$\times$128) to accommodate different instances. Next, we feed the resized crop sequence into v2e \cite{hu2021v2e}, a video-to-event converter for realistic events simulation, to obtain the corresponding raw events. Finally, we follow prior works \cite{rebecq2019high,Mei_2023_CVPR,Delbruck_2023_CVPR} to convert the raw events into event frames. By the above steps, we can obtain the intensity sequence and corresponding events sequence as well as emotion label triplets, covering diverse examples.}

\subsection{Dataset Statistical Analysis}
\mhy{We further present dataset statistics in in \autoref{tab_eff}.
It can be seen that (i) our DSEE dataset provides a substantial edge with its abundance of sequences, frames, and subjects, exceeding the capacity of current alternatives and enabling robust and generalizable research endeavors; (ii) our DSEE also covers participants with a wider range of age and more race numbers, providing more diversity for training and evaluating the model; and (iii) both real and synthetic events data are included in our DSEE, facilitating both the synthetic-data-based and real-data-based research and experiments, as well as the further exploration of synthetic-to-real transfer.}
\section{Experiments}\label{sec:exp}

\setlength{\tabcolsep}{2.6pt}
\begin{table*}[htbp]
	\small
	\centering
    \scalebox{1.0}{
	\begin{tabular}{l|c|ccccccc|cccc|cc}
	\toprule
  \multirow{2}{*}{Methods} & \multirow{2}{*}{\makecell[c]{input\\region}} & \multicolumn{7}{c|}{Accuracy for different emotions} & \multicolumn{4}{c|}{Accuracy for different lightings} & \multicolumn{2}{c}{Metrics (\%)} \\
	 &  &  Ha & Sa & An  &  Di &  Su & Fe & Ne & \makebox[0.045\textwidth][c]{Nor} & \makebox[0.045\textwidth][c]{Over} & \makebox[0.045\textwidth][c]{Low} & \makebox[0.045\textwidth][c]{HDR} & WAR $\uparrow$ & UAR $\uparrow$ \\
	\midrule
     Resnet18 + LSTM  & Face & 57.8 & 86.0 & 64.9 & 46.5 & 9.2 & 81.6 & 59.8 & 57.9 & 60.4 & 53.9 & 52.5 & 56.3 & 58.0 \\
    Resnet50 + GRU & Face & 27.9 & 38.0 & 49.7 & 44.5 & 6.9 & 70.0 & 5.6 & 43.0 & 35.7 & 28.9 & 32.8 & 35.2 & 34.7 \\
	 3D Resnet18~\cite{hara2018can} & Face & 54.8 & 45.4 & 67.7 & 23.8 & 37.2 & 42.8 & 81.6 & 51.9 & 51.4 & 44.8 & 47.8 & 49.1 & 50.5  \\
R(2+1)D~\cite{tran2018closer} & Face & 63.6 & 45.5 & 65.7 & 27.8 & 33.3 & 37.9 & 86.6 & 54.3 & 50.3 & 44.4 & 49.3 & 49.7 & 51.5 \\
Former DFER~\cite{zhao2021former} & Face & 81.5 & 75.2 & 85.8 & 59.4 & 39.3 & 50.8 & 78.6 & 70.1 & 65.4 & 66.2 & 61.1 & 65.8 & 67.2\\
Former DFER w/o pre-train & Face & 44.1 & 65.2 &  46.0 & 66.5 & 28.0  & 50.3  & 36.1 & 47.0 & 51.9  & 45.6 & 47.2 & 48.0 & 48.0  \\
    \midrule
Eyemotion~\cite{hickson2019eyemotion} & Eye & 74.3 & 85.5 & 79.5 & 74.3 & 69.1 & 79.2 & \underline{94.5} & 79.0 & 81.8 & 81.5 & 72.5 & 78.8 & 79.5  \\
Eyemotion w/o pre-train & Eye & 79.6 & 85.7 & 81.2 & 71.2 & 54.7 & 71.6 & \textbf{96.4} & 77.8 & 75.9 & 79.8 & 69.7 & 75.9 & 77.2  \\
EMO~\cite{wu2020emo} & Eye & 75.0 & 75.1 & 70.2 & 48.1 & 37.5 & 54.1 & 82.8 & 61.8 & 62.8 & 60.1 & 69.6 & 63.1 & 63.3  \\
EMO w/o pre-train & Eye& 79.6 & 85.7 & 81.2 & 71.2 & 54.7 & 71.6 & \textbf{96.4} & 77.8 & 75.9 & 79.8 & 69.7 & 75.9 & 77.2\\
    SEEN~\cite{zhang2023blink} & Eye & 85.0 & 89.9 & \underline{92.2} & 76.7 & 72.1 & 87.7 & 85.2& \underline{83.3} & \underline{85.6} & 80.8 & \textbf{84.8} & 83.6 & 84.1  \\
    \midrule  
    MSKD (Ours) \textit{teacher network} & Eye & \textbf{85.6} & \textbf{91.7} & \textbf{92.3} & \underline{79.0} & \textbf{79.4} &  \underline{88.0} & 90.3 & \textbf{84.4} & \textbf{89.1} & \textbf{88.3} & 82.7 & \textbf{86.2} & \textbf{86.6}\\
    MSKD (Ours) \textit{student network} & Eye &  \underline{83.2} & \underline{90.6} & 89.8 & \textbf{79.3} & \underline{72.2} & \textbf{89.4} & 86.3 & \underline{83.3} & \underline{85.6} & \underline{83.9} & \underline{83.0} & \underline{84.0} & \underline{84.4} \\
    \bottomrule
	\end{tabular}}
 \caption{\mhy{Quantitative comparison against state-of-the-arts. All methods are retrained and tested on the \textbf{SEE} dataset. The abbreviations are defined as Ha $\rightarrow$ Happiness; Sa $\rightarrow$ Sadness; An $\rightarrow$ Anger; Di $\rightarrow$ Disgust; Su $\rightarrow$ Surprise; Fe $\rightarrow$ Fear; Ne $\rightarrow$ Neutrality; Nor $\rightarrow$ Normal; Over $\rightarrow$ Overexposure; Low $\rightarrow$ Low-Light. The first and second best results are highlighted in \textbf{bold} and \underline{underline}, respectively.}}
	\label{tab:overall_see}
\end{table*}

\subsection{Experimental Settings}
\paragraph{Implementation Details.}
\mhy{We implement our method in PyTorch \cite{NEURIPS2019_9015} and perform the experiments on a server with Intel(R) Xeon(R) Gold 6240R @ 2.40GHz CPU and NVIDIA GeForce RTX 4090 GPU. The training of our method can be divided into two stages. First, we pretrain the ANN-SNN-hybrid teacher network with both intensity frame and events data as input, for 180 epochs with a batch size of 32. The training is optimized via stochastic gradient descent (SGD) with a momentum of $0.9$ and a weight decay of 0.001. The learning rate is initialized as 0.015 and decayed by a factor of 0.94 after each epoch. Second, our multimodality synergistic knowledge distillation strategy is adopted for the training of unimodal SNN student network. Except for the loss function, other training configurations for the student network are identical to those of the teacher network.}

\paragraph{Evaluation Dataset and Metrics.}
\mhy{We evaluate the effectiveness of method
on both the existing event-based single-eye emotion recognition dataset SEE \cite{zhang2023blink} as well as our newly constructed DSEE dataset. We adopt two widely used metrics for quantitatively assessing the emotion classification performance: Unweighted Average Recall (UAR) and Weighted Average Recall (WAR). UAR reflects the average accuracy of different emotion classes without considering instances per class, while WAR indicates the accuracy of overall emotions~\cite{2011Cross}. Formally,
\begin{align}
    UAR &= \frac{1}{N_c}\sum_{i=1}^{N_c}\frac{TP_i}{TP_i + FN_i}, \\
    WAR &= \frac{TP + FN}{TP + TN + FP + FN},  
\end{align}
where $N_c$ is the total number of emotion classes; $TP$ and $FP$ are true positive and false positive, respectively; $TN$ and $FN$ are true negative and false negative, respectively. For UAR and WAR, a higher value indicates a better performance.}



\subsection{Comparison to State-of-the-arts} 
\mhy{\autoref{tab:overall_see} reports the comparison results of our proposed MSKD approach against state-of-the-art methods on SEE dataset \cite{zhang2023blink}. From the results, we observe that: (i) eye-based emotion recognition methods are generally superior to facial-based ones. This reflects that the eyes serve as a crucial region for emotional expression as it contains rich information about subtle nuances in facial expressions. Focusing on this specific region enables more precise feature extraction, enhancing the discriminative power of the algorithm. Furthermore, eye-based methods are more robust to different light conditions. The reason behind this is that unlike the entire face, which may exhibit variations in shadowing and contrast under different lighting conditions, the eyes are less susceptible to these variations. The relatively small size of the eyes and their position in the middle of the face makes them less prone to extreme illumination changes, resulting in increased algorithmic robustness; (ii) our teacher network achieves the best overall performance. For example, it outperforms the existing state-of-the-art method SEEN \cite{zhang2023blink} by 2.6\% and 2.5\% in terms of WAR and UAR, respectively. This demonstrates the superior capability of the teacher network, laying a strong foundation for the subsequent training of the student network; and (iii) the proposed multimodality synergistic knowledge distillation strategy enables the student network to achieve comparable results to SEEN \cite{zhang2023blink} despite the absence of event data. This demonstrates the effectiveness of our strategy in knowledge transfer and opens doors for efficient model training in modality-scarce scenarios. We also validate the effectiveness of our method on our newly constructed DSEE dataset. Our proposed method also demonstrates competitive performance on our DSEE dataset, as evidenced by the top two rankings of our teacher and student networks in \autoref{tab:overall_dsee}. Notably, our student network surpasses SEEN \cite{zhang2023blink} by a large margin in both WAR (3.4\%) and UAR (3.1\%). This superiority remains consistent across diverse emotion categories and varied lighting conditions, demonstrating the robust capability of our approach for emotion recognition.}


\setlength{\tabcolsep}{2pt}
\begin{table*}[tbp]
	\small
	\centering
    \scalebox{1.0}{
	\begin{tabular}{l|c|ccccccc|cccc|cc}
	\toprule
  \multirow{2}{*}{Methods} & \multirow{2}{*}{\makecell[c]{input\\region}} & \multicolumn{7}{c|}{Accuracy for different emotions} & \multicolumn{4}{c|}{Accuracy for different lightings} & \multicolumn{2}{c}{Metrics (\%)} \\
	 &  &  Ha & Sa & An  &  Di &  Su & Fe & Ne & \makebox[0.045\textwidth][c]{Nor} & \makebox[0.045\textwidth][c]{Over} & \makebox[0.045\textwidth][c]{Low} & \makebox[0.045\textwidth][c]{HDR} & WAR $\uparrow$ & UAR $\uparrow$ \\
	\midrule
     Resnet18 + LSTM  & Face & 72.3 & 61.9 & 78.3 & 76.8 & 69.9 & \textbf{66.9} & 84.8 & 71.3 & 73.1 & 69.7 & 79.6 & 72.2 & 73.0 \\
    Resnet50 + GRU & Face & 66.3 & 62.2 & 78.3 & 75.8 & 68.5 & 65.7 & 84.7 & 69.9 & 70.5 & 69.3 & 78.4 & 71.6 & 66.3 \\
	 3D Resnet18~\cite{hara2018can} & Face & 62.7 & 56.5 & 58.9 & 50.9 & 51.5 & 33.5 & 62.6 & 54.9 & 55.6 & 44.2 & 56.5 & 53.3 & 53.8 \\
R(2+1)D~\cite{tran2018closer} & Face & 52.1& 53.5& 27.1&62.0 &54.7 &29.8 &35.6& 49.1& 45.9& 36.6&44.4& 45.8 & 45.0\\
Former DFER~\cite{zhao2021former} & Face & 71.1 & 54.8 & 67.2 & 64.3 & 45.2 & 42.7 & 82.6 & 58.1 & 64.2 & 60.7 & 58.8 & 59.7 & 61.1 \\
Former DFER w/o pre-train & Face & 63.1 & 50.7 & 49.9 & 48.1 & 40.8 & 42.0 & 62.8 & 50.3 & 50.7 & 47.3 & 53.5 & 50.2 & 51.1  \\
    \midrule
Eyemotion~\cite{hickson2019eyemotion} & Eye & 72.3 & 66.4 & 76.9 & 74.9 & 70.4 & 64.9 & \underline{85.8} & 71.3 & 
72.1 & 69.3 & 82.0 & 72.3 & 73.1    \\
Eyemotion w/o pre-train & Eye & 71.0 & 68.2 & 75.0 & 74.8 & 66.8 & 67.2 & \textbf{86.3} & 70.4 & 70.4 & 
70.1 & 83.8 &71.8&72.7    \\
EMO~\cite{wu2020emo} & Eye & 73.7 & 59.2 & 70.8 & 74.2 & 61.6 & 61.7 & 80.2 & 67.0 & 68.2 & 65.5 & 76.5 & 68.0 & 68.8   \\
EMO w/o pre-train & Eye & 68.3 & 64.4 & 70.7 & 73.7 & 62.9 & 58.6 & 82.5 & 67.4 & 67.0 & 63.2 & 78.6 & 67.8 & 68.7  \\
    SEEN~\cite{zhang2023blink} & Eye & 72.2 & 65.3 & 78.5 & 74.9 & 72.0 & 61.0 & 84.2 & 70.9 & 74.8 & 
69.8 & 75.5 & 71.9 & 72.6  \\
    \midrule
    MSKD (Ours) \textit{teacher network} & Eye &\textbf{80.1}&\textbf{73.7}&\textbf{83.8}&\underline{79.3}&\textbf{75.4}&\underline{66.8}&\textbf{86.3}&\textbf{76.1}&\textbf{78.6}&\textbf{75.0}&\textbf{86.2}& \textbf{77.4}&\textbf{77.9} \\
    MSKD (Ours) \textit{student network} & Eye &\underline{ 73.8} & \underline{71.9} & \underline{80.2} & \textbf{80.3} & \underline{73.7} & \textbf{66.9} & 82.8 & \underline{73.7} & \underline{76.2} & \underline{72.4} & \underline{85.9} & \underline{75.3}  &\underline{75.7}  \\
    \bottomrule
	\end{tabular}}
 \caption{\mhy{Quantitative comparison against state-of-the-arts. All methods are retrained and tested on the \textbf{DSEE} dataset. The abbreviations are defined as Ha $\rightarrow$ Happiness; Sa $\rightarrow$ Sadness; An $\rightarrow$ Anger; Di $\rightarrow$ Disgust; Su $\rightarrow$ Surprise; Fe $\rightarrow$ Fear; Ne $\rightarrow$ Neutrality; Nor $\rightarrow$ Normal; Over $\rightarrow$ Overexposure; Low $\rightarrow$ Low-Light. The first and second best results are highlighted in \textbf{bold} and \underline{underline}, respectively.}}
	\label{tab:overall_dsee}
\end{table*}


\subsection{Efficiency Analysis}

\mhy{The objective of this work is to develop an effective yet lightweight single-eye emotion recognition method to enhance its practical usability. Having established the effectiveness of our single-eye emotion recognition method, this section delves into a meticulous analysis of its efficacy along three key dimensions: computational complexity, inference speed, and energy cost. \autoref{tab:model_compare} shows that among the existing eye-based emotion recognition methods, Eyemotion \cite{hickson2019eyemotion} outperforms EMO \cite{wu2020emo} and SEEN \cite{zhang2023blink} in emotion recognition accuracy but lags in efficiency. Benefiting from the utilization of SNNs and weight-copy scheme, ANN-SNN-hybrid SEEN \cite{zhang2023blink} achieves comparable recognition accuracy as ANN-based Eyemotion \cite{hickson2019eyemotion} in a more efficient way. Our method takes advantage of multimodality synergistic knowledge distillation, achieving the best recognition performance and better computational efficiency than SEEN \cite{zhang2023blink}. In addition, we conduct the energy cost comparison in \autoref{tab:energy}. Energy estimations are predicated on \cite{horowitz20141}’s examination of 45 nm CMOS technology, as adopted in \cite{rathi2021diet,li2021differentiable}, in which SNN addition operations cost 0.9 $pJ$ whereas ANN MAC operations demand 4.6 $pJ$. From the results we can draw the conclusion that our method is much more energy efficient than existing methods, \textit{e.g.}, 108.52 and 10.76 times more energy efficient than Eyemotion \cite{hickson2019eyemotion} and SEEN \cite{zhang2023blink}, respectively.}


\begin{table}[tb]
	\centering
	\small
	\setlength{\tabcolsep}{5pt}{
		\begin{tabular}{l|c|c|c|c}
			\toprule
			Methods & WAR & FLOPs (G)& Params (M)& Time (ms)\\
			\midrule
			Eyemotion & 72.3 & 5.73 & 25.13 & 17.5\\
			\midrule
			EMO & 68.0 & 0.32 & \textbf{1.68} & 7.1\\
			\midrule
			SEEN & 71.9 & 0.95 & 6.08 & 7.2\\
            \midrule
			Ours & \textbf{75.3} & \textbf{0.27} & 4.04 & \textbf{6.1}\\
			\bottomrule
		\end{tabular}
        \caption{\mhy{Computational efficiency comparison of different eye-based emotion recognition methods.}}
		\label{tab:model_compare}}
\end{table}

\begin{table}[tb]
	\centering
	\small
	\setlength{\tabcolsep}{4pt}{
		\begin{tabular}{l|c|c|c|c}
			\toprule
			Methods & Eyemotion & EMO & SEEN & MSKD (Ours) \\
			\midrule
			Energy ($mJ$) & 823.69 & 46.00 & 81.64 & \textbf{7.59} \\
            \midrule
            Multiple to Ours & 108.52 & 6.06 & 10.76 & 1.00 \\
			\bottomrule
		\end{tabular}
  \caption{\mhy{Estimated energy efficiency comparison of different eye-based emotion recognition methods.}}
		\label{tab:energy}}
\end{table}


\subsection{Ablation Study}


\paragraph{Effectiveness of Hit / Temporal Consistency Knowledge Distillation.}
\mhy{\autoref{tab:ablation2} reports the results on the DSEE testing set when using different loss functions to train the student network. It can be seen that: (i) the temporal efficient training (TET) approach \cite{deng2022temporal} can benefit the training of SNN-based student network, \textit{i.e.}, \textit{b} is better than \textit{a}; (ii) both hit and temporal consistency knowledge distillation can help improve the performance (\textit{c} and \textit{d} perform better than \textit{b}); and (iii) the complementary nature of hit and temporal consistency knowledge distillation is evident in their synergistic interaction, which leads to demonstrably superior model performance (\textit{e} is better than \textit{c} and \textit{d}).}


\paragraph{Effects of different event data configurations.}
\mhy{We explored multiple ways to utilize event data as input for training our teacher network, aiming to determine the most effective approach. To enable recognition during any phase of the emotion, we adopted \cite{zhang2023blink}'s approach of using a uniformly distributed random starting point and corresponding sequence length for testing. Specifically, a start point is selected to ensure that the rest sequence is longer than the testing length which is defined as the total accumulation time of all included event frames, $x$, and a skip time between two adjacent event frames where all events are ignored, $y$, denoted as $ExSy$. We express both accumulation time and skip time as multiples of 1/30 s. This yields a testing length of $ExSy$ corresponding to $(x + (x-1)\times y)/30$ s. From \autoref{tab:ablation1}(\textit{A}-\textit{F}, \textit{M}), we observe that: (i) with four input event frames, longer skip time yields better performance (\textit{i.e.}, \textit{A}, \textit{B}, \textit{C}, and \textit{M} gets better in order). This shows that a longer testing length is beneficial as it contains more temporal information; and (ii) benefited from more temporal cues, with the same $3\times(1/30)$ s skip time, more event frames generate improved performance (\textit{D}, \textit{E}, \textit{F}, and \textit{M} gets better in order).}




\begin{table}[!t]
    \setlength{\tabcolsep}{6pt}	
    \centering
    \small
		\begin{tabular}{cl|cc}
        \toprule
            & Training Losses &  WAR $\uparrow$ & UAR $\uparrow$\\
            \midrule
            \textit{a} & $\mathcal{L}_{\text{Cls}}$ \textit{w/o} TET \cite{deng2022temporal} & 70.77  & 71.37 \\
			\textit{b} & $\mathcal{L}_{\text{Cls}}$ &71.90 &72.57 \\
			\textit{c} & $\mathcal{L}_{\text{Cls}}$ + $\mathcal{L}_\text{HCKD}$ & 73.83 &74.15\\
			\textit{d} & $\mathcal{L}_{\text{Cls}}$ + $\mathcal{L}_\text{TCKD}$&74.42 &74.88\\
			\textit{e} & $\mathcal{L}_{\text{Cls}}$ + $\mathcal{L}_\text{HCKD}$ + $\mathcal{L}_\text{TCKD}$ &\textbf{75.27} &\textbf{75.66}\\
        \bottomrule
		\end{tabular}
  \caption{\mhy{Ablation study on synergistic knowledge distillation.}}
    \label{tab:ablation2}
\end{table}

\paragraph{Influence of loss function in MSKD.} 
\mhy{There are multiple choices when measuring the distance between distributions in the knowledge distillation process, such as cross-entropy (CE), mean squared error (MSE), and Kullback-Leibler divergence (KLD). As can be seen from \autoref{tab:ablation1}(\textit{I}-\textit{M}), replacing MSE with CE or KLD in HCKD or TCKD leads to performance degradation (\textit{i.e.}, \textit{I}-\textit{L} are lower than \textit{M}). This indicates that MSE is a better choice for our synergistic knowledge distillation. Besides, removing the TET \cite{deng2022temporal} in classification loss $\mathcal{L}_{\text{Cls}}$ or temporal consistency knowledge distillation loss $\mathcal{L}_\text{TCKD}$ deteriorates the performance (\textit{G} and \textit{H} are worse than \textit{M}), showing the effectiveness of the temporal efficient training strategy.}



\begin{table}[tbp]
    \setlength{\tabcolsep}{8pt}
    \centering
    \small
    \scalebox{0.95}{
	\begin{tabular}{cl|cc}
	\toprule
    & Networks & WAR $\uparrow$ & UAR $\uparrow$   \\  
    \midrule
    $A$	&  $E4S0$ & 67.00 & 68.05 \\  
    $B$ & $E4S1$  & 72.21 & 72.85 \\ 
    $C$	&  $E4S2$ & 73.62 & 74.17 \\  
    $D$ & $E1S3$  & 62.83 & 63.58 \\ 
    $E$	&  $E2S3$ & 68.92 & 69.67 \\  
    $F$ & $E3S3$  & 72.22 & 72.77 \\ 
    \midrule
    $G$	& No TET for $\mathcal{L}_{\text{Cls}}$ & 74.79& 75.23  \\ 
    $H$	& No TET for $\mathcal{L}_{\text{TCKD}}$ & 74.32&74.77   \\ 
    \midrule
    $I$	&  $\mathcal{L}_{\text{HCKD}}$: MSE $\rightarrow$ CE  & 74.91&75.35   \\ 
    $J$	&  $\mathcal{L}_{\text{HCKD}}$: MSE $\rightarrow$ KLD  & 73.87&74.34 \\ 
    
    $K$ &   $\mathcal{L}_{\text{TCKD}}$: MSE $\rightarrow$ CE  & 74.31 & 74.71  \\ 
    $L$ &   $\mathcal{L}_{\text{TCKD}}$: MSE $\rightarrow$ KLD & 73.94 & 74.42  \\ 
    \midrule
    $M$	& Ours (\textit{E4S3}; \textit{w/} TET; MSE) & \textbf{75.27} & \textbf{75.66} \\ 
	\bottomrule
	\end{tabular}
	}	
 \caption{\mhy{Quantitative ablation results indicate that each key component in our MSKD contributes to the overall performance.}}
    \label{tab:ablation1}
\end{table}

\section{Conclusion}\label{sec:conclusion}
\mhy{In summary, our research introduces a pioneering approach to enhance single-eye emotion recognition in source-limited wearable devices. Drawing inspiration from apprenticeship learning, our multimodality synergistic knowledge distillation mechanism empowers a lightweight spiking neural network for more effective yet efficient recognition. Extensive validation demonstrates its significant improvement over state-of-the-art methods in both accuracy and efficiency. Furthermore, the establishment of a diverse single-eye emotion benchmark not only validates our method but also lays the foundation for continued exploration and innovation in the realm of single-eye-based methodologies. This work signifies a notable advancement in practical usability and user experience in the context of emotion recognition, opening new avenues for innovation and exploration in this dynamic field.}

\section*{Acknowledgements}
This work was supported in part by National Key Research and Development Program of China (2022ZD0210500), the National Natural Science Foundation of China under Grants 62332019/U21A20491, and the Distinguished Young Scholars Funding of Dalian (No. 2022RJ01). Yang Wang was supported by a Chinese Scholarship Council (CSC) grant. This work is also supported by IAF, A*STAR, SOITEC, NXP, National University of Singapore under FD-fAbrICS: Joint Lab for FD-SOI Always-on Intelligent \& Connected Systems (Award I2001E0053). Haiyang Mei and Mike Zheng Shou did not receive any funding for this work.

\bibliographystyle{named}
\bibliography{paper}

\end{document}